\documentclass[11pt,a4paper]{article}
\usepackage[hyperref]{eacl2021}
\usepackage{times}
\usepackage{latexsym}

\usepackage{amsmath}
\usepackage{amssymb}
\usepackage{float}
\usepackage{graphicx}
\usepackage{lipsum}
\usepackage{multirow}
\usepackage{blindtext}
\usepackage{wrapfig}
\usepackage{color}
\usepackage{blindtext}
\usepackage{caption}
\usepackage{graphicx,subcaption}
\usepackage{tabularx}
\usepackage{microtype}
\usepackage{footnote}
\usepackage{amssymb}
\usepackage{enumerate}
\usepackage{enumitem}

\usepackage{microtype}
\makeatletter
\def\thickhline{%
  \noalign{\ifnum0=`}\fi\hrule \@height \thickarrayrulewidth \futurelet
   \reserved@a\@xthickhline}
\def\@xthickhline{\ifx\reserved@a\thickhline
               \vskip\doublerulesep
               \vskip-\thickarrayrulewidth
             \fi
      \ifnum0=`{\fi}}
\makeatother

\newlength{\thickarrayrulewidth}
\aclfinalcopy 

\title{On the Importance of Local Information in Transformer Based Models}

\author{Madhura Pande, Aakriti Budhraja, Preksha Nema \\ \textbf{Pratyush Kumar}, \textbf{Mitesh M. Khapra} \\
\\
Department of Computer Science and Engineering \\ 
Robert Bosch Centre for Data Science and AI (RBC-DSAI) \\ Indian Institute of Technology Madras, Chennai, India \\ 
\texttt{\{mpande,abudhra,preksha,pratyush,miteshk\}}@cse.iitm.ac.in}

\date{}

\begin{document}
\maketitle
\begin{abstract}
The self-attention module is a key component of Transformer-based models, wherein each token pays attention to every other token. 
Recent studies have shown that these heads exhibit syntactic, semantic, or local behaviour. Some studies have also identified promise in restricting this attention to be \textit{local}, i.e., a token attending to other tokens only in a small neighbourhood around it. 
However, no conclusive evidence exists that such local attention alone is sufficient to achieve high accuracy on multiple NLP tasks. 
In this work, we systematically analyse the role of locality information in learnt models and contrast it with the role of syntactic information. 
More specifically, we first do a sensitivity analysis and show that, at every layer, the representation of a token is much more sensitive to tokens in a small neighborhood around it than to tokens which are syntactically related to it. 
We then define an attention bias metric to determine whether a head pays more attention to local tokens or to syntactically related tokens. We show that a larger fraction of heads have a locality bias as compared to a syntactic bias. 
Having established the importance of local attention heads, we train and evaluate models where varying fractions of the attention heads are constrained to be local. Such models would be more efficient as they would have fewer computations in the attention layer. 
We evaluate these models on 4 GLUE datasets (QQP, SST-2, MRPC, QNLI) and 2 MT datasets (En-De, En-Ru) and clearly demonstrate that such constrained models have comparable performance to the unconstrained models. 
Through this systematic evaluation we establish that attention in Transformer-based models can be constrained to be local without affecting performance.
\end{abstract}

\section{Introduction}

Transformer based models \cite{vaswani2017attention,devlin2018bert} have produced state-of-the-art results in different NMT \cite{sutskever2014sequence} and NLU tasks \cite{wang2018glue}. 
A key component of these models is the multi-head self attention network which computes a contextual representation for each token by considering all other tokens in the sentence. 
The success of this multi-head self-attention network has motivated several studies in interpreting the role of each attention head, beginning with the encoder. 
Relevant questions in such studies \cite{voita2019analyzing,clark2019does,michel2019sixteen}, are: What does a particular head attend to? Is there a semantic explanation of such attention? Do all heads contribute equally? If not, which heads are higher contributors? One common finding of these studies is that a few heads which are either syntactic or local are more important than other heads. A syntactic head is one which pays more attention to tokens which are syntactically related to a given token whereas a local head is one which  pays more attention to close-by tokens (upto $k$ tokens on either side of the given token). These studies thus suggest that syntactic and local information is important.

In this work, we dig deeper into the role of locality information in Transformer-based models. In particular, we first perform a sensitivity analysis and show that, at every layer, the representation of a token is more sensitive to local tokens as compared to other tokens. Further, we show that this sensitivity to local tokens is higher than that to syntactically related distant tokens. This suggests that at every layer, local information is more important than non-local or even syntactic information. But, what about attention heads? Do they show any bias by paying more attention to certain types of tokens, either local or syntactic? To check this, we define an attention bias score which is the fraction of attention paid to specific tokens (local or syntactic). Using this score we find that a larger number of heads have a locality bias whereas only few heads have a syntactic bias. These results establish the primacy of local information. 

Motivated by the above findings, we pose the following question: ``Can all attention heads be constrained to be local'', 
i.e., can Transformer-based models have high accuracy if we introduce a model bias of only local attention heads? 
If true, then this would reduce the parameters and computations of Transformer networks.
To answer the above question, we propose a modified Transformer where we combine the standard attention heads with local attention heads to varying degrees in each layer of the encoder. 
In particular, we evaluate configurations in which the attention heads in each layer are explicitly made local by multiplying with a binary mask to zero out the attention on all tokens farther than a distance of $k$.
For example, one head attends to only the previous token and gives zero attention to all other tokens, while another head attends to the next two tokens and gives zero attention to all other tokens and so on. 
To further study possible variations in local attention, we analyse configurations in which different attention heads share the same parameters (key and query matrices) but make different choices on the positions of their attention, i.e., the binary masks.
For example, a layer would share one set of parameters but use different binary masks for each attention head.


We extensively evaluate these modified Transformer configurations on different Machine Translation and Natural Language Understanding tasks. 
In particular, we evaluate on the English-German(EN-DE) and English-Russian(EN-RU) MT tasks using the WMT'14 and Newstest2018 datasets with Transformer model with a modified encoder. 
We report that most configurations have comparable performance with the baseline (maximum BLEU score drop of 0.24 and 0.52 for EN-DE and EN-RU respectively). 
In particular, the configuration that uses only local attention heads on all layers has a BLEU score drop of 0.14 in the EN-RU task and a BLEU score increase of 0.11 in the EN-DE task. 
For an extreme configuration where in each layer all heads have the same set of parameters but different sieves, the BLEU score drops by only 0.27 in the EN-RU task and increases by 0.3 in the EN-DE task. 
Similarly, for four NLU tasks (QQP, SST-2, MRPC, QNLI) from the GLUE benchmark we train modified BERT models. 
We observe only a 2-3\% drop using a constrained network as compared to an unconstrained network. 
The extreme configuration where all layers share parameters sees a larger drop in accuracy, much of which is recovered when un-tying the parameters of the last two layers.
These results show that local attention in most layers is sufficient to obtain high accuracy in Transformer-based models.

\section{Background: Transformer Based Models}
In this work, we analyse transformer based models used for NMT \cite{vaswani2017attention} and NLU \cite{devlin2018bert}.
Several configurations of these models have been proposed, of which we use (i) the Transformer$_{BASE}$ configuration as defined in \cite{vaswani2017attention} for NMT and (ii) the BERT$_{BASE}$ configuration as defined in \cite{devlin2018bert} for NLU. 
The Transformer$_{BASE}$ model contains an encoder and a decoder each having 6 layers. Each of these layers has 8 self-attention heads. 
Similarly, the BERT$_{BASE}$ model contains 12 encoder layers and each of these layers has 12 self-attention heads. 
For all our analysis, we will focus only on the encoder.

For a specific layer, let the input be given as $\mathbf{X} \in \mathbb{R}^{T\times d_v}$, where $T$ is the number of input tokens and $d_v$ is the size of the embedding used to represent the tokens. 
An attention head transforms input $\mathbf{X}$ into three distinct vectors namely the Key, Query, and Value with learnt matrices $\mathbf{W}^k, \mathbf{W}^q, \mathbf{W}^v$, respectively, as $\mathbf{K} = \mathbf{X}\mathbf{W}^k, \quad  \mathbf{Q} = \mathbf{X} \mathbf{W}^q, \quad \mathbf{V} = \mathbf{X} \mathbf{W}^v$, where  $\mathbf{K,Q,V} \in \mathbb{R}^{T\times d_l}$, where $d_l$ is the size of the internal representation within the head. 
After $\mathbf{K, Q, V}$ are computed, the output of attention is given as:
\begin{equation}
    \text{Attention}(\mathbf{Q, K, V}) = \text{softmax}\left(\dfrac{\mathbf{QK^\intercal}}{\sqrt{d_l}}\right)\mathbf{V}
    \label{eq:attn}
\end{equation}
The result of the Attention operation is a $T \times d_l$ matrix where the output at each token is a linear combination of the values of all tokens with weights given by the softmax computation. 
For multi-headed self-attention, the above operation is repeated across $H$ heads and the result concatenated and linearly combined with a weight matrix $\mathbf{W}^o \in \mathbb{R}^{d_l H\times d_v}$ as:
\begin{align}
    \mathbf{Y} &= \mbox{Concat}(\mbox{head}_1, \ldots , \mbox{head}_H)\mathbf{W}^o, \label{eq:attn_output} \\ 
    \mbox{head}_h &= \mbox{Attention}(\mathbf{XW}_h^q, \mathbf{XW}_h^k, \mathbf{XW}_h^v) \\
    & = \mbox{softmax}\left(\dfrac{\mathbf{XW}_h^q(\mathbf{XW}_h^k)^\intercal}{\sqrt{d_l}}\right)\mathbf{XW}_h^v 
    \label{eq:attn_single}
\end{align}

The self-attention module is followed by a residual connection which sums $\mathbf{X}$ and $\mathbf{Y}$.
This sum is then the input to a feed-forward network of 2 linear transformations with ReLU non-linear activation function.
The output of this feed-forward network is the input to the next layer. 
Thus, the attention heads at each layer are characterised by the parameters: $\mathbf{W}_h^k, \mathbf{W}_h^q, \mathbf{W}_h^v, \mathbf{W}_h^o$.
For the Transformer$_{BASE}$ configuration the dimensions are given by $d_v = 512, d_l = 64, H = 8$. 

\begin{figure*}
   \begin{tabular}{cccc}
\includegraphics[width=1.4in]{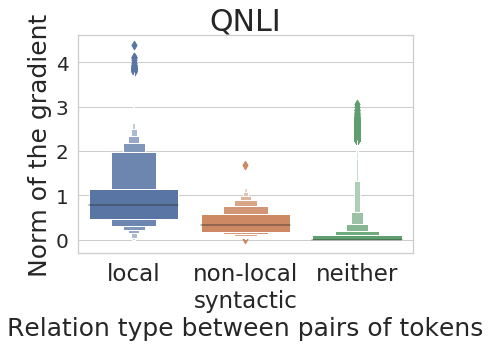} &
\includegraphics[width=1.4in]{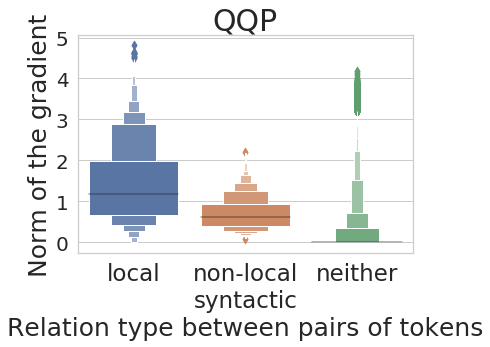} &
\includegraphics[width=1.4in]{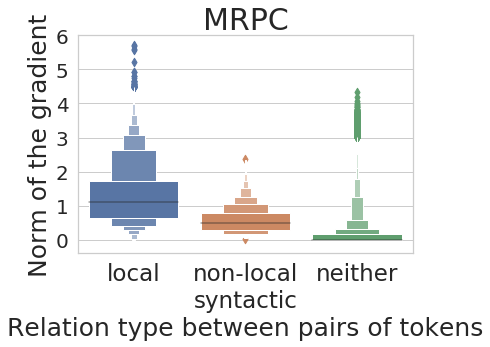}&
\includegraphics[width=1.4in]{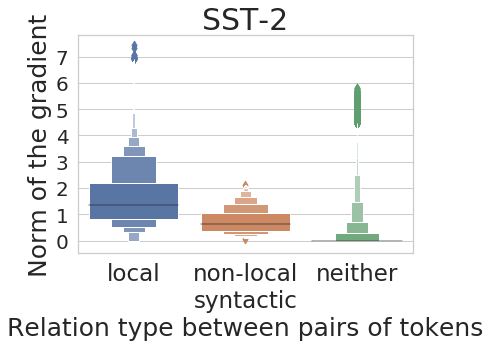} \\
   \end{tabular}
    \caption{Distribution of the $\gamma_{local}$ (blue), $\gamma_{syntactic}$ (orange), and $\gamma_{neither}$ (green) for the the four GLUE tasks. Norm gradient of tokens are larger for local dependency indicating the relative importance of local information.}
    \label{table:gradient_analysis}
\end{figure*}

\section{Analysis of Locality in Transformer based models}
In this section, we analyse trained Transformer-based models to check the importance of local information in these models. To do this, we first train Transformer$_{BASE}$ models for En-Ru and En-De translation tasks (using WMT'14). Our trained models achieve a BLEU score of $29.09$ and $30.01$ for En-Ru and En-De respectively on the Newstest2018 dataset. Note that these BLEU scores are within 0.6 BLEU points of the performance of Transformer$_{BASE}$ reported elsewhere \cite{voita2019analyzing}. Similarly, we pre-train and fine-tune the BERT$_{BASE}$ model for four NLU tasks (QQP, SST-2, MRPC, QNLI) from the GLUE benchmark \cite{wang2018glue}. Again, the performance of our trained models on these tasks is within 1\% of the performance of BERT$_{BASE}$ reported elsewhere \cite{devlin2018bert}. We mention these numbers to convince the reader that the models achieve SOTA results and hence are reliable for further analysis. In the following sub-sections, we systematically analyse the role of local information.

\subsection{Sensitivity analysis}
We first study the gradients of the output of multi-headed self-attention as a function of the input tokens. 
Specifically, for each layer of the encoder, we compute a matrix $\beta \in \mathbb{R}^{T \times T}$, where $\beta[i, j]$ denotes the norm of the partial derivative of $\mathbf{Y}[i]$ w.r.t. the input $\mathbf{X}[j]$. Here, $\mathbf{Y}[i]$ is the representation of the $i$-th token at the output of multi-headed self attention (see Equation \ref{eq:attn_output}) and $\mathbf{X}[j]$ is the representation of the $j$-th token at the input of multi-headed self attention. For every token $i$, we then define three sets $L_i, S_i$ and $U_i$. Here, $L_i = [i - 2, i + 2]$ is the set of all indices which are within a window of 2 from $i$. Similarly, $S_i$ contains the indices of all tokens which are in a syntactic relation with the token at location $i$ but are not in the local neighborhood of $i$. This definition of $S_i$ enables us to study those syntactic relations which are not already covered by local relations. Lastly, $U_i$ contains the indices of all tokens which are neither in $L_i$ nor in $S_i$ (i.e., it contains the indices of tokens which are neither locally nor syntactically related to the token at location $i$). 
We can then compute the following values corresponding to the sensitivity of the output to local, syntactic and unrelated tokens:
\vspace{-0.5em}
\begin{align}
    \gamma_{local} &= \mbox{avg}_{i \in [1,T]} \left(\mbox{avg}_{j \in L_i}\beta[i, j]\right) \label{eq:gamma_local}\\
    \gamma_{syntactic} &= \mbox{avg}_{i \in [1,T]} \left(\mbox{avg}_{j \in S_i}\beta[i, j]\right) \label{eq:gamma_syntactic}\\
    \gamma_{unrelated} &= \mbox{avg}_{i \in [1,T]} \left(\mbox{avg}_{j \in U_i}\beta[i, j]\right)
    \label{eq:gamma_unrelated}
\end{align}

In the definition of $\gamma_{local}$, the quantity inside the bracket on the RHS is the average norm of the gradient of the $i$-th output token w.r.t. its locally related input tokens. This is further averaged over all the $T$ tokens in the sentence to obtain $\gamma_{local}$ which gives us the average sensitivity of the output tokens to locally related tokens. Similarly, $\gamma_{syntactic}$ and $\gamma_{unrelated}$ give us the average sensitivity of the output tokens to syntactically related tokens and unrelated tokens respectively. 

In Figure \ref{table:gradient_analysis}, we show the distribution of $\gamma_{local}$, $\gamma_{syntactic}$ and $\gamma_{unrelated}$ for 100 randomly sampled sentences each from the test sets of the 4 GLUE tasks( QQP, SST-2, MPRC and QNLI). 
Across all the tasks, we observe that $\gamma_{local}$ is higher than $\gamma_{syntactic}$ which in turn is higher than $\gamma_{unrelated}$. This confirms that across all the tasks the output of each layer is most sensitive to the inputs that are local. Further, the sensitivity to unrelated tokens is negligible.

\subsection{Attention Bias Metric}
Given the $i$-th token and a subset of other tokens at indices $\mathbf{i'}$, we can ask the question: When computing a representation of the $i$-th token how attentive is a head to tokens at $\mathbf{i'}$.
A disproportionately high fraction of attention would imply that the head has an \textit{attentive bias} from $i$-token towards tokens at indices $\mathbf{i'}$.
We can capture this with an {\em attention bias metric} defined as the ratio of two averages.
In the numerator, we have the average attention paid by the head from the $i$-th token to tokens in $\mathbf{i'}$. 
In the denominator, we have the average attention paid by the head from the $i$-th token to all tokens in the sentence. 
Higher this ratio, higher is the attentive bias towards $\mathbf{i'}$.
Mathematically, this attention bias metric is given as:
\begin{equation}
    a_h(i, \mathbf{i'}) = \left(\dfrac{\sum_{j \in \mathbf{i'}} \alpha_h[i, j]}{|\mathbf{i'}|}\right) \bigg/ \left(\dfrac{\sum_{j \in \mathbf{T}} \alpha_h[i, j]}{|\mathbf{T}|}\right),
\end{equation}
where $\alpha_h[i, j]$ denotes the attention paid in head $h$ from $i$-th token to $j$-th token, and $\mathbf{T}$ is the set of all tokens in an input sentence. 

A head that is randomly initialised would pay roughly equal attention to all heads and would have an attentive bias metric close to 1 for every choice of $\mathbf{t'}$.
If upon learning, a head becomes more selective, then its attentive bias for specific subsets would increase.
We would like to compute such attentive bias scores for specific choices of tokens - namely local tokens and syntactic tokens. 
In particular, we can compute the attentive bias of a head to local tokens by choosing $\mathbf{i'} = L_i$.  
We can then average such terms across all tokens of a sentence, and then across all sentences to compute a single \textit{locality bias score} for each head.
A larger locality bias score would imply that that head selectively attends to neighbouring tokens.
Similarly, by choosing $\mathbf{i'} = S_i$, i.e., the set of all tokens to which $i$ has a syntactic relation but are not local, we can compute the \textit{non-local syntactic bias score} for each head.

\begin{figure*}
    \centering
    \includegraphics[width=6.2in]{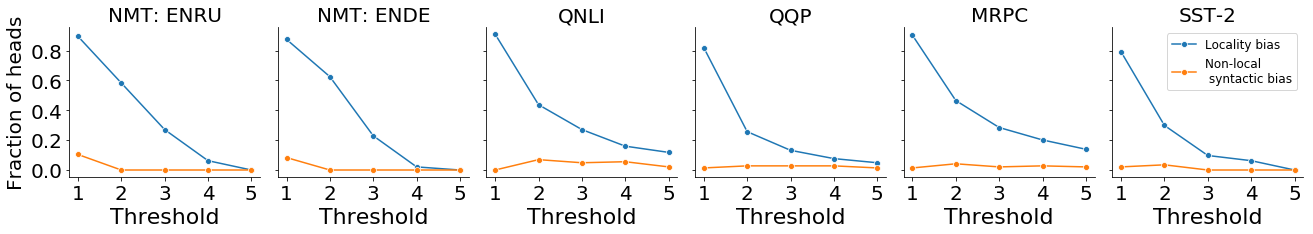}
    \caption{Fraction of heads with locality bias score greater than a threshold (blue) and non-syntactic bias score greater than a threshold (orange) for different threshold values across 2 NMT and 4 GLUE tasks.}
    \label{fig:fractional_plots}
\end{figure*}

\begin{figure*}
  \begin{tabular}{@{\hskip 0in}c@{\hskip 0in}c@{\hskip 0in}c@{\hskip 0in}c@{\hskip 0in}c@{\hskip 0in}c@{\hskip 3in}}
   \includegraphics[width=1.1in]{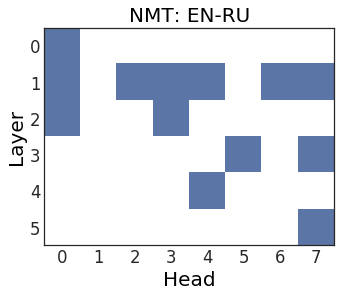} & 
   \includegraphics[width=1.1in]{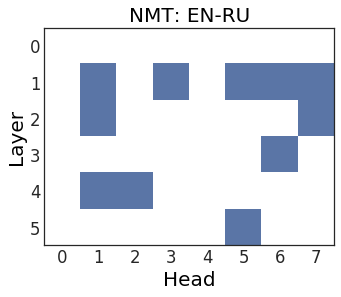} & 
    \includegraphics[width=1in]{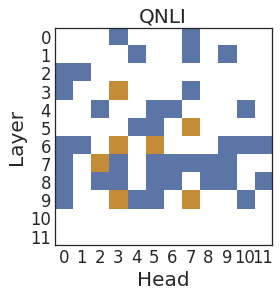} &
     \includegraphics[width=1in]{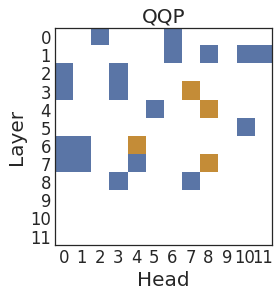} & 
      \includegraphics[width=1in]{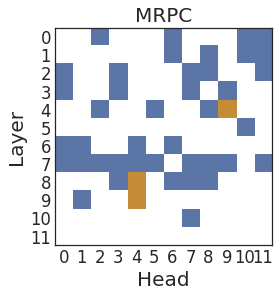} &
       \includegraphics[width=1in]{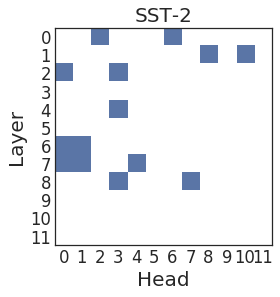} \\
     \end{tabular}

    \caption{Set of heads which have a locality bias score greater than 3 (blue) and non-local syntactic bias score greater than 3 (orange) for 2 NMT and 4 GLUE tasks. Large number of heads with higher locality bias score indicates the importance of local information.}
    \label{fig:heatmaps}
\end{figure*}

We compute the locality and non-local syntactic bias scores for 1,000 sentences each from the test sets of the tasks that we considered (QQP, SST-2, MPRC and QNLI).
In Figure \ref{fig:fractional_plots}, we show the fraction of heads which have a locality bias score greater than a given threshold which is varied from 1 through 5.
Note again that a randomly initialised head would be expected to have a locality bias score of 1.
Higher the threshold, the larger is the expected attentive bias, and thus fewer are the heads which have such sharp local bias. 
In 4 out of the 6 tasks, more than a quarter of heads have a locality bias greater than a high threshold of 3, thereby indicating a significant role of local information.
We also plot the fraction of heads which have a non-local syntactic bias greater than a given threshold. 
Clearly, the fraction of heads with non-local syntactic bias is significantly smaller than those with locality bias at all values of the threshold. 
To analyse this further, we zoom into these results for the specific choice of 3 as threshold.
In Figure \ref{fig:heatmaps}, we mark layer-wise the individual heads which have a locality bias greater than 3 across each of the 6 tasks (shown in blue).
We see that the heads with local bias are spread across layers, though later layers have fewer such heads.
We also mark layer-wise the heads which have a non-local syntactic bias, and find that there are far fewer such heads (shown in orange).

\subsection*{Summary}
In this section, with the help of sensitivity analysis and attention bias score we showed that local information plays a very important role in transformer based models. 
A conclusive way of establishing this would be to evaluate the Transformer with the model bias of only local attention. 
Differently stated, if we were to enforce that all the cells of Figure \ref{fig:heatmaps} are purple (i.e., have a value of 1), how accurate would Transformers be? This will be discussed in the remainder of this paper.

\section{Transformers With Only Local Attention}

In this section, we propose our modification to the Transformer network to support only local attention in the encoder. 
We do this in two steps. 
In the first step, we introduce masks to create multiple templates of local attention. 
In the second step, we propose parameter sharing to further constrain local attention across attention heads.
\subsection{Applying Masks to Attention Heads}
A simple way to constrain specific heads to local attention is by multiplying a binary mask to the computed attention matrix $\alpha_h$. 
We thus introduce the following {\em mask} operation in each layer of the encoder:
\begin{align} \nonumber 
\tilde{\alpha_h} = \text{mask}_h \odot \mbox{softmax}\left(\dfrac{(\mathbf{X}\mathbf{W}_h^q)(\mathbf{X}\mathbf{W}_h^k)^T}{\sqrt{d_l}}\right)
\end{align}
\begin{align}
    = \text{mask}_h \odot \alpha_h,
\label{eq:mask}
\end{align}
where $\text{mask}_h$ is a binary matrix of size $T \times T$, and $\alpha_h$ is the $T \times T$ matrix denoting the pair-wise attention on head $h$.
The output $Y$  (Equation \ref{eq:attn_output}) is then computed with the modified attention map $\tilde{\alpha_h}$.

To model local attention, the mask vectors should have 1's close to the main diagonal, and 0's elsewhere.
We define a family of such masks to provide diversity in training heads that have local attention: \texttt{prev-1}, \texttt{prev-2}, \texttt{next-1}, \texttt{next-2}, \texttt{band-1}, \texttt{band-2} and \texttt{identity}. A \texttt{prev-k} mask would be a $T \times T$ matrix whose $(i,j)$-th entry will be 1 if $j - i = k$ and 0 otherwise. In other words, it would be an identity matrix whose columns are left shifted by $k$. Similarly, a \texttt{next-k} mask would be an identity matrix whose columns are right shifted by $k$. Lastly, a \texttt{band-k} mask would be a $T \times T$ matrix whose $(i,j)$-th entry will be 1 if $ i - k \leq j \leq i + k$ and 0 otherwise. In other words, it will mask out the attention to all tokens which are not within a window of $k$ around a given token.
If different attention heads in a layer have different masks, it affords the network the ability to learn to prioritise across them. 
For instance, for a period (`.') token the prev1 and prev2 masks may be more appropriate than the next1 and next2 masks.

Notice that in our modified Transformer, attention is first computed in the form of $\alpha_h$, and then the mask is applied.
Hence, the values in $\tilde{\alpha}_h$ would retain the coefficients as computed with softmax over all tokens, but only in the positions where the mask is 1.
A more efficient implementation can avoid softmax computation across all tokens, and thereby save on training and inference time.
These savings can be significant because (a) the number of input tokens can be very large as opposed to our local window of only 5 tokens, and (b) the softmax operation is computationally expensive.
The focus in the current work is to show that such masking achieves results comparable to the baseline. 
More efficient implementation is a direction of future work.

The application of the mask operation of Equation \ref{eq:mask} can be viewed as a {\em regularisation} on the Transformer model: 
We bias the model to constrain the encoding where each layer has a small \textit{receptive field}.
Longer range relations can still be encoded due to the stacking of multiple layers (typically 6 for Transformer encoder and 12 for BERT).
In the experimental section, we evaluate the effect of such regularisation on accuracy.

\subsection{Parameter Sharing Across Heads}
The formulation of Equation \ref{eq:mask} motivates the following question: 
Can we decouple the learning of $\alpha_h$ from the choice of mask?
More specifically: Can each attention head be characterised by a separate mask, but share the same $\alpha_h$ with other heads?
Sharing the same $\alpha_h$ is equivalent to parameter sharing of the weight matrices $\mathbf{W}^q, \mathbf{W}^k$ across different heads.

With parameter sharing, there are multiple design choices: 
What is the mask to choose for each head? Which heads should share parameters? Should attention heads across multiple layers share parameters?
Clearly, the larger the number of heads which share parameters, the greater the regularisation on the encoder.
In the experimental section, we evaluate the effect of these choices on the accuracy of the model. 

Like in the case of masking, parameter sharing provides an opportunity to make Transformers more efficient.
Shared parameters reduce the model size: Shared weight matrices are stored just once and $\alpha_h$ is computed just once.
In the experimental section, we report reduction in model size, but leave the analysis of compute efficiency to a future work.

\section{Results and Discussion}

In this section, we report experimental results of training and evaluating Transformer models with the introduction of the model bias of local attention. 

\subsection{Experimental Setup}
We train the Transformer$_{BASE}$ model on WMT'14 English-Russian [EN-RU] and English-German [EN-DE] datasets and evaluate it using Newstest2014. 
All the training experiments are run from scratch and were run until convergence on single V100 and 1080TI GPUs. 
Batch size for EN-RU was set to 25K while that for EN-DE was 32K. 
The high batch sizes are required for training \cite{popel2018training}, and were obtained by accumulating gradients.
We used Adam optimizer with $\beta_1$ = 0.9, $\beta_2$ = 0.997 and $\epsilon$ = 10$^{-9}$. 
We varied the learning rate according to the formula described in \cite{vaswani2017attention} with \textit{warmup\_steps} = 16k.
Similarly, we pre-train the BERT$_{BASE}$ model using the English Wikipedia corpus and a subset of of the Project Gutenberg corpus released by \newcite{lahiri:2014:SRW}. We  pre-trained the model for around 300k steps with a sequence length of 128 and another 1k steps with the sequence length of 512 with batch size of 2K \cite{devlin2018bert}. We then tune this model individually for four NLU tasks from the GLUE benchmark. For each task, we used the standard train and test splits for fine-tuning and evaluation. We used the recommended setting of hyperparameters(\cite{devlin2018bert}) with batch sizes chosen among \{32, 128\} and learning rates among \{1e-4, 2e-4\}. Warm-up was set to 10K steps and LAMB optimizer was used.

In the next sub-sections, we first present detailed analysis and results for the two NMT tasks. Based on the insights from these results, we then perform a smaller set of experiments on the GLUE tasks. 

\begin{table*}
    \centering
    \begin{tabular}{ccccccc}
    \hspace{-1.5cm}\includegraphics[width=2cm,height=1.35cm]{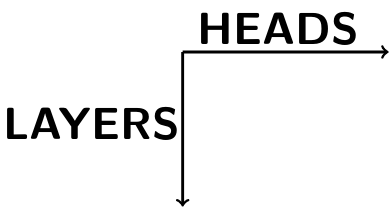} &
    \includegraphics[width=1.8cm,height=1cm]{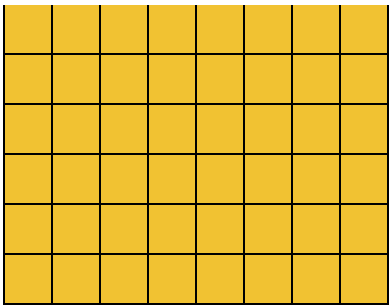} &
    \includegraphics[width=1.8cm,height=1cm]{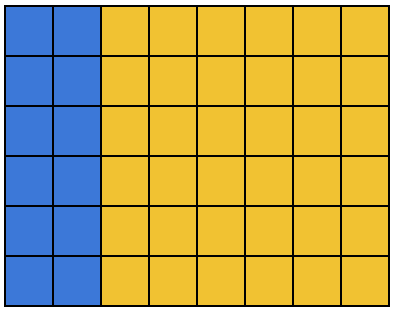} &
    \includegraphics[width=1.8cm,height=1cm]{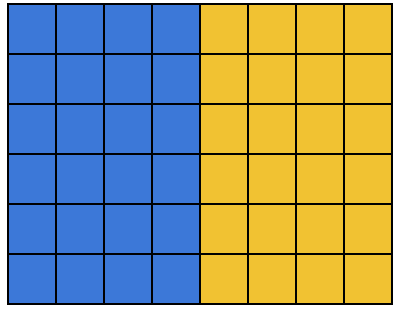} &
    \includegraphics[width=1.8cm,height=1cm]{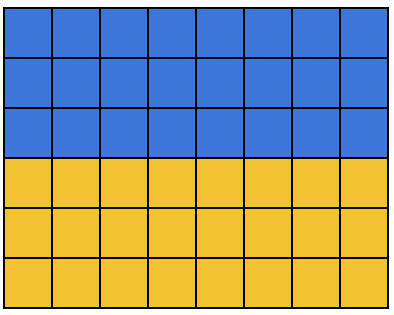} &
    \includegraphics[width=1.8cm,height=1cm]{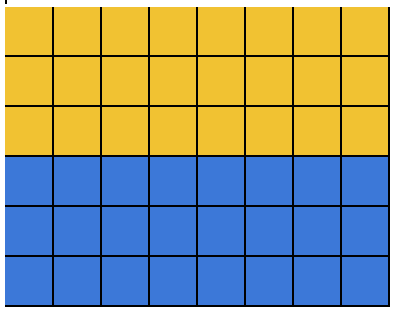} &
    \includegraphics[width=1.8cm,height=1cm]{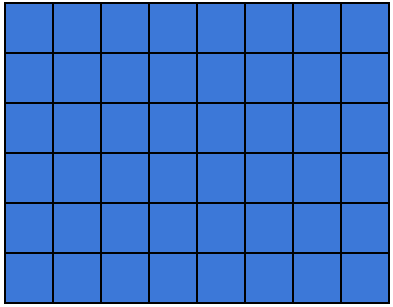}\\
     & (a) & (b) & (c) & (d) & (e) & (f)\\
    \includegraphics[width=2cm,height=1.4cm]{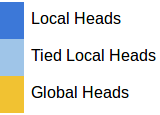} &
    \includegraphics[width=1.8cm,height=1cm]{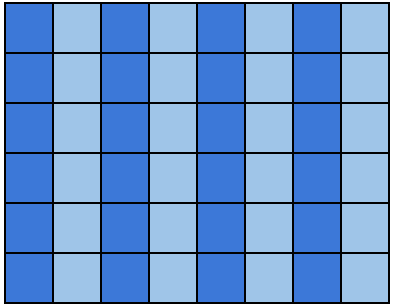} &
    \includegraphics[width=1.8cm,height=1cm]{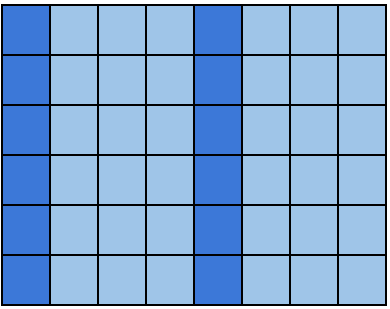} &
    \includegraphics[width=1.8cm,height=1cm]{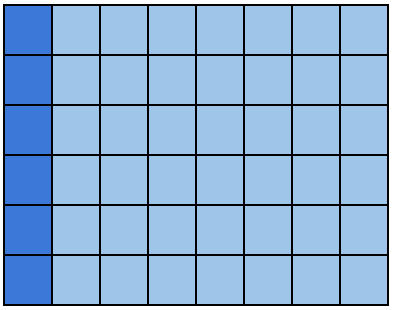} &
    \includegraphics[width=1.8cm,height=1cm]{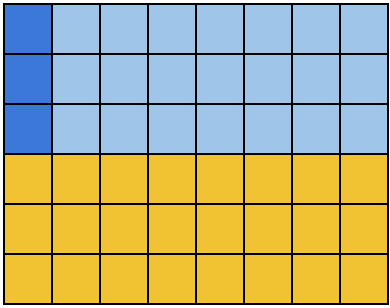} & 
    \includegraphics[width=1.8cm,height=1cm]{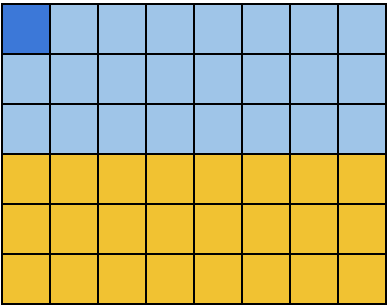} &
    \includegraphics[width=1.8cm,height=1cm]{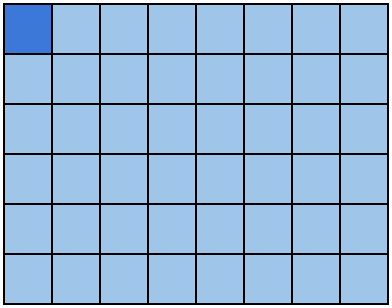}\\
    & (g) & (h) & (i) & (j) & (k) & (l)\\
    \end{tabular}
    \captionof{figure}{Graphical representation of different configurations: (a) Base model, (b)-(f): Heads with local attention positioned in various layers. (g)-(l) Heads with local attention with parameter sharing.}
    \label{fig:configs}
\end{table*}

\subsection{Intermixing Global and Local Attention Heads}
In this first experiment, we evaluate multiple configurations which combine attention heads with global (standard) and local (proposed) attention. 
The configurations are defined by two choices: (a) how many and which attention heads are to be constrained to local attention, and (b) which masks (as defined in Equation \ref{eq:mask}) are to be used to characterise the local attention.
For the former, we make 6 different choices of laying out global and local attention heads as shown pictorially in Figure \ref{fig:configs}(a)-(f). 
These configurations go from the baseline model as shown in (a) to the extreme case of replacing all heads with local attention heads as shown in (f).
Configurations (b) and (c) explore the effect of increasing the number of local attention heads per layer.
Configurations (d) and (e) explore the choice of distributing local attention across early or later layers.
The choice of masks depends on the number of local attention heads; we make the following choices: 

\noindent-~\textbf{2LocHeads\_All6 (config b)}: In each layer, 2 heads attend locally. The masks are band1 and band2. \\
\noindent-~\textbf{4LocHeads\_All6 (config c)}: In each layer, 4 heads attend locally. The masks are band1 and band2, repeated twice each. \\
\noindent-~\textbf{8LocHeads\_First3 (config d) and 8LocHeads\_Last3 (config e)}: In 3 layers, all attention heads are local with all masks used once and identity mask repeated twice. \\
\noindent-~\textbf{8LocHeads\_All6 (config f)}: All layers have all attention heads as local with all masks used once and identity mask repeated twice.

For each of the 6 configurations, we train the Transformer models from scratch. 
We report the BLEU scores on the two NMT tasks in Table \ref{table:untied_results}.
Across both tasks and all configurations the maximum drop in BLEU score is only 0.14.
Indeed, for the case of constraining all attention heads to local attention, the BLEU score drops by 0.14 for EN-RU, but increases for EN-DE. 
On the EN-DE task, BLEU scores for all modified configurations are higher than the baseline.
This demonstrates that local attention is sufficient. 
The complex relationships between input tokens are being successfully encoded by attention heads regularised to only attend locally.

\begin{table}
\centering
	\begin{minipage}[c]{0.45\textwidth}
	\resizebox{1\textwidth}{!}{
		\begin{tabular}{p{3cm}p{0.75cm}p{1.2cm}p{1.2cm}}
        \thickhline
        \multirow{2}{*}{Experiment} & \multirow{2}{*}{Fig \ref{fig:configs}} & \multicolumn{2}{c}{BLEU Score}  \\ 
         \cline{3-4}
          & & EN-RU & EN-DE \\ 
        \thickhline
            \textbf{Baseline} & \textbf{(a)} & \textbf{29.09} & \textbf{30.01}  \\ 
        \hline
        
            2LocHeads\_All6 & (b) & 29.37 & 30.78 \\
            4LocHeads\_All6 & (c) & 29.34 & 30.83 \\
            8LocHeads\_First3  & (d) & 28.96 & 30.6   \\
            8LocHeads\_Last3 & (e) & 28.99 & 30.51 \\
            8LocHeads\_All6 & (f) & 28.95 & 30.12  \\ 
        \thickhline
\end{tabular} }
\caption{BLEU scores of various arrangements of local heads within a layer. xLocHeads\_y indicates x local heads in each of the y encoder layers.}
\label{table:untied_results}
\end{minipage}
\hfill%
\begin{minipage}[c]{0.5\textwidth}
\resizebox{0.9\textwidth}{!}{
\begin{tabular}{ccccc}
\thickhline
\multirow{2}{*}{Experiment} & \multirow{2}{*}{Fig \ref{fig:configs}} & \multirow{2}{1cm}{Attn params} &  \multicolumn{2}{c}{BLEU Score}  \\ 
 \cline{4-5} &
  & & EN-RU & EN-DE \\ 
\thickhline
    \textbf{Baseline} & \textbf{(a)} & \textbf{6.29M} & \textbf{29.09} & \textbf{30.01}  \\ 
\hline
    4LocHeads\_4TiedLoc\_All6 & (g) & 4.71M &  28.83 & 30.24 \\
    2LocHeads\_6TiedLoc\_All6 & (h) & 3.93M &  28.57 & 30.12 \\
    1LocHead\_7TiedLoc\_All6 & (i) & 3.53M & 28.82 & 30.31 \\
    1LocHead\_7TiedLoc\_First3 & (j) & 4.91M & 29.43  & 30.51 \\
\thickhline
\end{tabular}}
\caption{BLEU score for different arrangements of local heads with \textbf{layer-wise} parameter tieing. xLocHeads\_yTiedLoc\_z indicates x local heads, y tied-local heads in each of the z layers.}
\label{table:tied_within_layer}
\end{minipage}
\end{table}

\subsection{\textbf{Parameter Sharing Within Layer}} 
We now constrain the attention heads further by sharing the parameters $\mathbf{W}^k, \mathbf{W}^q$.
The 4 configurations used in these experiments are shown in Figure \ref{fig:configs}(g)-(j). 
In these configurations, the dark blue cells denote local attention heads which have a unique set of parameters ($\mathbf{W}^q, \mathbf{W}^k$). 
The light blue cells denote local attention heads which share the parameters from one of the attention heads represented by the dark blue cells.
The different configurations we evaluate are detailed below.

\noindent-~\textbf{4LocHeads\_4TiedLoc\_All6 (config g)}: All attention heads are local.
    In each layer, 4 attention heads have unique parameters, while the other heads share the parameters from their previous head. 
    For each of the 4 pairs of attention heads sharing the same parameters, identity and band2 are the two masks.\\
\noindent-~\textbf{2LocHeads\_6TiedLoc\_All6 (config h)}: All attention heads are local.
    In each layer, 2 attention heads (head 0 and 4) have unique parameters, while the other heads share the parameters.
    Thus, heads 1, 2, and 3 use $\alpha_0$ computed in head 0, while heads 5, 6, and 7 use $\alpha_4$ computed in head 4.
    The masks used are identity, band2, prev1, next1 which is repeated twice.\\
\noindent-~\textbf{1LocHead\_7TiedLoc\_All6 (config i)}: All attention heads are local. 
    Each layer has one set of unique parameters shared across all other heads. 
    All masks are used once and identity mask is repeated twice. \\
\noindent-~\textbf{1LocHead\_7TiedLoc\_First3 (config j)}: Last 3 layers have global attention, while first 3 layers have local attention. 
    Each of the first three layers has a single head with unique parameters shared with all heads of that layer. 
    In the first three layers, all masks are used once and identity mask repeated twice.

The results of this experiment are shown in Table \ref{table:tied_within_layer}.
Along with the BLEU scores for EN-RU and EN-DE tasks, we show the number of attention parameters in the encoder as a whole. 
The maximum drop in the BLEU score across all configurations for the EN-RU tasks is 0.52. 
For the EN-DE configuration, each of the four configurations had a higher BLEU score than the baseline.
For both tasks, config i provides a good trade-off between decrease in parameters in the encoder and the BLEU score.
    
\subsection{\textbf{Parameter Sharing Across Layers}}
In these final set of configurations, we evaluate the sharing of parameters across multiple layers.
We consider two configurations shown in Figure \ref{fig:configs} (k)-(l), which are described below.\\
\noindent-~\textbf{Half tied (config k)}: 
    The first three layers have local attention with all heads sharing the same parameters, while the last three layers are global attention. 
    In each of the first three layers all 7 masks are used, with the identity mask used twice.\\
\noindent-~\textbf{Fully tied (config l)}: 
    This is the extreme case where a single set of parameters is used across all layers which are constrained to have local attention heads. 
    In each layer, all 7 masks are used, with the identity mask used twice.

The results are shown in  Table \ref{table:across_layer_tied}. 
Even with the maximum parameter sharing in the Fully Tied configuration, BLEU scores drop by 0.1 and 0.24 respectively for EN-RU and EN-DE tasks. 
Thus, even with half the attention parameters, the impact on the model accuracy is limited.
\begin{table}
\begin{minipage}[c]{0.45\textwidth}
	\resizebox{1\textwidth}{!}{
\begin{tabular}{ccccc}
\thickhline
\multirow{2}{*}{Experiment} &  \multirow{2}{*}{Fig \ref{fig:configs}} & \multirow{2}{1cm}{Attn params} & \multicolumn{2}{c}{BLEU Score}  \\ 
 \cline{4-5} & &
  & EN-RU & EN-DE \\ 
\thickhline
    \textbf{Baseline} & (a) & 6.29M  &\textbf{29.09} & \textbf{30.01}  \\ 
\hline
    Half Tied & (k) & 4.78M & 29.24 & 30.08 \\
    Fully Tied & (l) & 3.21M & 28.99 & 29.77  \\ 
\thickhline
\end{tabular} 
}
\caption{\label{font-table} BLEU score with a \textbf{single} local head with its parameters tied across layers: (i) with all heads of first 3 layers(first row) and ,(ii) with all heads of all layers(second row).}
\label{table:across_layer_tied}
\end{minipage}
\hfill%
\begin{minipage}[c]{0.5\textwidth}
	\resizebox{1\textwidth}{!}{
\begin{tabular}{ccccc}
\thickhline
Model &  QQP & SST-2 & MNLI & QNLI \\
\thickhline
    $BERT_{BASE}$ & 90.9 & 91.8 & 83.7 & 90.8  \\ 
\hline
    All\_Band2\_Untied & 90.05 & 89.75 &  80.6 & 88.3 \\
    All\_Band6\_Untied & 90.46 & 90.65 &  81.15 & 89 \\
    All\_Band6\_AllTied & 86.5 & 87.8 & 67.6  & 72.1 \\
    All\_Band6\_132Tied & 88.8 & 91.3 & 74.3 & 83.2 \\
    All\_Band6\_120Tied & 90.4 & 91.16 &  80.7 & 88.45 \\
\thickhline
\end{tabular} }
\caption{Results on GLUE Dev set for various configurations of the BERT model.}
\label{table:BERT_Results}
\end{minipage}

\end{table}

\subsection{BERT Results}
The above results reported on the two NMT tasks clearly suggest that the performance does not drop much even when (i) we restrict all the heads to be local and (ii) a large number of local heads in the network share their parameters. To check if these findings are indeed consistent across other tasks we perform a focused set of experiments on four GLUE tasks. The idea is to check if the above observations hold for these tasks also. We report these results in Table \ref{table:BERT_Results}. 
The first row reports the results for the standard BERT$_{BASE}$ model for comparison. The second and third row corresponds to the case when band2 and band6 masks are applied to all the 144 heads (12 layers of 12 heads each) in the network and the attention parameters are not shared. The last three rows correspond to parameter sharing. The fourth row corresponds to the case when band6 masks are applied to all the 144 heads and parameters are shared across all layers. The next two rows correspond to relaxing the parameter sharing to exclude the last layer (132 tied) and the last two layers (120 tied). 

We observe that in the case of untied models, constraining the local attention does not affect accuracy significantly. 
However, when tying all parameters the accuracy drops significantly, unlike in the case of the Transformer model.
This is reasonable: Unlike in the Transformer model with a decoder module, the last layer of the BERT model is required to output the final result and thus suffers from parameter sharing. 
In conformance with this, we observe that untying the last one and two layers (132 and 120 tied heads) recovers the accuracy with untied models.
Thus, remarkably we achieve high accuracy with a modified BERT model where all heads are constrained to be local and 10 of the 12 layers share parameters.

\section{Related work}
Our work has relation to existing works on (i) local attention in the context of seq2seq and self attention networks (ii) analysis of attention heads in transformers 
and (iii) paramater sharing in NMT. We review some papers in each of these categories below.

The idea of local attention in the context of NMT was first proposed by \newcite{luong2015effective} who limited the attention to 
\newcite{luong2015effective}
a subset of source words at a time. By doing so they were able to get improved performance for En-De translation. Similarly, \newcite{tjandra-etal-2017-local} and \newcite{inproceedings} used local monotonic attention and time-restricted local attention respectively to improve the performance of automatic speech recognition. \newcite{DBLP:conf/interspeech/Junfeng} assumed attention to be a time-moving Gaussian window with the parameters of the Gaussian controlling the locality of the attention mechanism. Even in the context of self attention networks, some works \cite{yang-etal-2018-modeling,9003949,Guo2019GaussianTA} have used a Gaussian bias to model local attention. The idea is to introduce a learnable Gaussian mask centered around the current word which weakens the attention on distant words. On the other hand, \newcite{Dai2020} propose a multiple positional self attention network which weakens or avoids the attention on distant words. There are also some works \cite{lioutas2020timeaware,DBLP:conf/iclr/WuFBDA19,DBLP:conf/naacl/YangWWCT19} which use lightweight convolutions with smaller receptive fields to capture only local information which computing token representations. Similarly,  \newcite{DBLP:conf/iclr/ShenZL0Z18} and \newcite{DBLP:conf/iclr/WuLLLH20} use a combination of local and long range attention to model different relationships between tokens.

Next, we review some papers which focus on analysing transformer based models. 
For example, \cite{voita2019analyzing} analyse the encoder of a transformer based NMT model and conclude that the important heads focus on \textit{positional}, \textit{syntactic} and \textit{rare-word} relations. Similarly, \cite{clark2019does} and \cite{vig2019visualizing} analyse BERT using attention-based probing classifiers and visualization tools. These and other studies \cite{DBLP:journals/corr/abs-1911-12246,DBLP:journals/corr/abs-1911-03898} conclude that attention heads capture positional, syntactic and/or semantic information. In contrast, our analysis emphasises on the importance of \textit{local} information.

One of the ideas proposed in this work is to share paramaters across different attention heads. The idea of parameter sharing is of course quite popular in Deep Learning and it has also been explored extensively in the context of NMT \cite{dong2015multi,firat2017multi,sachan2018parameter,xia2019tied}.
Even in the context of transformer based models, \cite{lan2019albert} have shown that BERT can be made much leaner by sharing parameters.

\section{Conclusion}

Through analysis of gradients and the proposed attention bias metrics we quantified the large importance of local information in Transformer-based models.
We then trained Transformer-based models that are constrained to only local attention and observed comparable accuracy on two NMT and four GLUE tasks.
Our results confirm that local attention suffice to learn accurate Transformer-based models.
We also showed that this restriction towards local attention can be combined with parameter sharing while still preserving high accuracy. 
These empirical results motivate the creation of efficient Transformer-based models that only pay attention locally. 



\bibliographystyle{acl_natbib}
\bibliography{main}

\end{document}